\documentclass[runningheads]{llncs}
\usepackage{graphicx}
\usepackage{algorithm}
\usepackage{algorithmic}
\usepackage{comment}

\usepackage{amsmath,amssymb} 
\usepackage{tikz}
\usepackage{comment}
\usepackage{amsmath,amssymb} 

\usepackage{color}
\usepackage{hyperref}

\begin{document}
\pagestyle{headings}
\mainmatter
\def\ECCVSubNumber{173}  

\title{Off-Policy Reinforcement Learning for Efficient and Effective GAN Architecture Search} 

\titlerunning{E$^2$GAN: Off-policy RL for GAN Architecture Search}

\author{Yuan Tian \inst{1}$^*$ \and
Qin Wang \inst{1}$^*$ \and
Zhiwu Huang \inst{1}\and
Wen Li \inst{2}\and
Dengxin Dai \inst{1}\and\\
Minghao Yang \inst{3} \and
Jun Wang \inst{4}\and
Olga Fink \inst{1}
}
\authorrunning{Y. Tian et al.}  
\institute{ETH Z\"urich\\
\email{\{yutian, qwang, ofink\}@ethz.ch} \email{\{zhiwu.huang, dai\}@vision.ee.ethz.ch} \and
UESTC \quad\quad\quad\quad\quad\; \inst{3} Navinfo Europe \quad\quad\;\inst{4} University College London \\
\email{liwenbnu@gmail.com} \,
\email{minghao.yang@navinfo.eu} \,
\email{junwang@cs.ucl.ac.uk}}

{\let\thefootnote\relax\footnotetext{$^*$ Equal contribution.}}
\maketitle
\begin{abstract}
In this paper, we introduce a new reinforcement learning (RL) based neural architecture search (NAS) methodology for effective and efficient generative adversarial network (GAN) architecture search. The key idea is to formulate the GAN architecture search problem as a Markov decision process (MDP) for smoother architecture sampling, which enables a more effective RL-based search algorithm by targeting the potential global optimal architecture. To improve efficiency, we exploit an off-policy GAN architecture search algorithm that makes efficient use of the samples generated by previous policies.
Evaluation on two standard benchmark datasets (i.e., CIFAR-10 and STL-10) demonstrates that the proposed method is able to discover highly competitive architectures for generally better image generation results with a  considerably reduced computational burden: 7 GPU hours. Our code is available at \href{https://github.com/Yuantian013/E2GAN}{https://github.com/Yuantian013/E2GAN}.
\end{abstract}

\section{Introduction}
Generative adversarial networks~(GANs) have been successfully applied to a wide range of generation tasks, including image generation~\cite{goodfellow2014generative,brock2018large,bao2017cvae,wang2018high,guo2019auto}, text to image synthesis~\cite{reed2016generative,zhang2017stackgan,park2019semantic} and image translation~\cite{isola2017image,choi2018stargan}, to name a few. To further improve the generation quality, several extensions and further developments have been proposed, ranging from regularization terms~\cite{brock2016neural,gulrajani2017improved}, progressive training strategy~\cite{karras2017progressive}, utilizing attention mechanism~\cite{xu2018attngan,zhang2018self}, and to new architectures~\cite{karras2019style,brock2018large}.

While designing favorable neural architectures of GANs has made great success, it typically requires a large amount of time, effort, and domain expertise. For instance, several state-of-the-art GANs~\cite{karras2019style,brock2018large} design appreciably complex generator or discriminator backbones for better generating high-resolution images. To alleviate the network engineering pain, an efficient automated architecture searching framework for GAN is highly needed. On the other hand, Neural architecture search~(NAS) has been applied and proved effective in discriminative tasks such as image classification~\cite{krizhevsky2012imagenet} and segmentation~\cite{liu2019auto}. Encouraged by this,  AGAN~\cite{wang2019agan} and AutoGAN~\cite{gong2019autogan} have introduced neural architecture search methods for GAN based on reinforcement learning (RL), thereby enabling a significant speedup of architecture searching process.

Similar to the other architecture search tasks (image classification, image segmentation), recently proposed RL-based GAN architecture search method AGAN~\cite{wang2019agan} optimized the entire architecture. Since the same policy might sample different architectures, it is likely to suffer from noisy gradients and a high variance, which potentially further harms the policy update stability. To circumvent this issue, multi-level architecture search
(MLAS) has been used in AutoGAN ~\cite{wang2019agan}, and a progressive optimization formulation is used. However, because optimization is based on the best performance of the current architecture level, this progressive formulation potentially leads to a local minimum solution.

To overcome these drawbacks, we reformulate the GAN architecture search problem as a Markov decision process~(MDP). The new formulation is partly inspired by the human-designed Progressive GAN~\cite{karras2017progressive}, which has shown to improve generation quality progressively in intermediate outputs of each architecture cell. In our new formulation, a sequence of decisions will be made during the entire architecture design process, which allows state-based sampling and thus alleviates the variance. In addition, as we will show later in the paper, by using a carefully designed reward, this new formulation also allows us to target effective global optimization over the entire architecture.

More importantly, the MDP formulation can better facilitate off-policy RL training to improve data efficiency. The previously proposed RL-based GAN architecture search methods~\cite{gong2019autogan,wang2019agan} are based on on-policy RL, leading to limited data efficiency that results in considerably long training time. 
Specifically, on-policy RL approach generally requires frequent sampling of a batch of architectures generated by current policy to update the policy.
Moreover, new samples are required to be collected for each gradient step, while the previous batches are directly disposed. This quickly becomes very expensive as the number of gradient steps and samples increases with the complexity of the task, especially in the architecture search tasks. by comparison, off-policy reinforcement learning algorithms make use of past experience such that the RL agents are enabled to learn more efficiently. This has been proven to be effective in other RL tasks, including legged locomotion~\cite{lillicrap2015continuous} and complex video games~\cite{mnih2015human}. However, exploiting off-policy data for GAN architecture search poses new challenges. Training the policy network inevitably becomes unstable by using off-policy data, because these training samples are systematically different from the on-policy ones. This presents a great challenge to the stability and convergence of the algorithm~\cite{bhatnagar2009convergent}. Our proposed MDP formulation can make a difference here. By allowing state-based sampling, the new formulation alleviates this instability, and better supports the off-policy strategy.

The contributions of this paper are two-fold:

\begin{enumerate}
   \item We reformulate the problem of neural architecture search for GAN as an MDP for smoother architecture sampling, which enables a more effective RL-based search algorithm and potentially more global optimization.
  \item We propose an efficient and effective off-policy RL NAS framework for GAN architecture search (E$^2$GAN), which is 6 times faster than existing RL-based GAN search approaches with competitive performance. 
\end{enumerate}

We conduct a variety of experiments to validate the effectiveness of E$^2$GAN. Our discovered architectures yield better results compared to RL-based competitors. E$^2$GAN is efficient, as it is able to to find a highly competitive model within \textbf{7 GPU hours}.

\section{Related Work}

\textbf{Reinforcement Learning}
Recent progress in model-free reinforcement learning (RL)~\cite{sutton1992reinforcement} has fostered promising results in many interesting tasks ranging from gaming~\cite{mnih2013playing,silver2014deterministic}, to planning and control problems~\cite{hwangbo2019learning,kumar2016learning,xie2019iterative,han2019h_,chao2020real,han2020actor} and even up to the AutoML~\cite{zoph2016neural,pham2018efficient,liu2018progressive}. However, model-free deep RL methods are notoriously expensive in terms of their sample complexity. One reason of the poor sample efficiency is the use of on-policy reinforcement learning algorithms, such as trust region policy optimization (TRPO)~\cite{schulman2015trust}, proximal policy optimization(PPO)~\cite{schulman2017proximal}  or REINFORCE~\cite{williams1992simple}. On-policy learning algorithms \textbf{require new samples generated by the current policy for each gradient step}. On the contrary, off-policy algorithms aim to reuse past experience. Recent developments of the off-policy reinforcement learning algorithms, such as soft Actor-Critic (SAC)~\cite{haarnoja2018soft}, have demonstrated substantial improvements in both performance and sample efficiency in previous on-policy methods.

\textbf{Neural architecture search}
Neural architecture search methods aim to automatically search for a good neural architecture for various tasks, such as image classification~\cite{krizhevsky2012imagenet} and segmentation~\cite{liu2019auto}, in order to ease the burden of hand-crafted design of dedicated architectures for specific tasks. Several approaches have been proposed to tackle the NAS problem. Zoph and Le~\cite{zoph2016neural} proposed a reinforcement learning-based method that trains an RNN controller to design the neural network~\cite{zoph2016neural}. Guo et al.~\cite{guo2019nat} exploited a novel graph convolutional neural networks for policy learning in reinforcement learning. Further successfully introduced approaches include evolutionary algorithm based methods\cite{real2017large}, differentiable methods~\cite{liu2018darts} and one-shot learning methods~\cite{brock2017smash,liu2018darts}. Early works of RL-based NAS algorithms~\cite{xie2018snas,pham2018efficient,zoph2016neural,liu2018progressive} proposed to optimize the entire trajectory (i.e., the entire neural architecture). To the best of our knowledge, most of the previously proposed RL-based NAS algorithms used on-policy RL algorithms, such as REINFORCE or PPO, except~\cite{zhong2018practical} which uses Q-learning algorithm for NAS, which is a value-based method and only supports discrete state space problems. For on-policy algorithms, since each update requires new data collected by the current policy and the reward is based on the internal neural network architecture training, the on-policy training of RL-based NAS  algorithms inevitably becomes computationally expensive.

\textbf{GAN Architecture Search}
Due to the specificities of GAN and their challenges, such as instability and mode collapse, the NAS approaches proposed for discriminative models cannot be directly transferred to the architecture search of GANs. Only recently, few approaches have been introduced tackling the specific challenges of the GAN architectures. Recently,  AutoGAN has introduced a neural architecture search for GANs based on reinforcement learning (RL), thereby enabling a significant speedup of the process of architecture selection~\cite{gong2019autogan}. 
The AutoGAN algorithm is based on on-policy reinforcement learning. The proposed multi-level architecture search (MLAS) aims at progressively finding well-performing GAN architectures and completes the task in around 2 GPU days. Similarly, AGAN~\cite{wang2019agan} uses reinforcement learning for generative architecture search in a larger search space. The computational cost for AGAN is comparably very expensive (1200 GPU days). In addition, AdversarialNAS~\cite{gao2019adversarialnas} and DEGAS~
\cite{doveh2019degas} adopted a different approach, i.e., differentiable searching strategy~\cite{liu2018darts}, for the GAN architecture search problem.

\section{Preliminary}
In this section, we briefly review the basic concepts and notations used in the following sections.

\subsection{Generative Adversarial Networks}
The training of GANs involves an adversarial competition between two players, a generator and a discriminator. The generator aims at generating realistic-looking images to `fool' its opponent. Meanwhile, the discriminator aims to distinguish whether an image is real or fake. This can be formulated as a min-max optimization problem:

\begin{equation}
\min_G\max_D \mathbb{E}_{x\sim p_{real}}[\log{D(x)}]  + \mathbb{E}_{z\sim p_{z}} [\log{(1-D(G(z)))}],
\end{equation}
where G and D are the generator and discriminator parametrized by neural networks. $z$ is sampled from random noise. $x$ are the real and $G(z)$  are the generated images. 

\subsection{Reinforcement Learning}
A Markov decision process (MDP) is a discrete-time stochastic control process. At each time step, the process is in some state $s$, and its associated decision-maker chooses an available action $a$. Given the action, the process moves into a new state $s'$ at the next step, and the agent receives a reward. 

An MDP could be described as a tuple 
($S,A,r,P,\rho$), where $S$ is the set of states that is able to precisely describe the current situation,
$A$ is the set of actions, $r (s,a)$ is the reward function, 
$P (s'|s,a)$ is the 
transition probability function, and 
$\rho (s)$ is the initial state 
distribution.

MDPs can be particularly useful for solving optimization problems via reinforcement learning. In a general reinforcement learning setup, an agent is trained to interact with the environment and get a reward from its interaction. The goal is to find a policy $\pi$ that maximizes the cumulative reward $J(\pi)$:

\begin{equation}
    J(\pi) = \mathbb{E}_{\tau \sim \rho_{\pi}}{\sum_{t=0}^{\infty} r(s_t, a_t)}
\label{eq:j}
\end{equation}

While the standard RL merely maximizes the expected cumulative rewards, the maximum entropy RL framework  considers a more general objective~\cite{ziebart2010modeling}, which favors stochastic policies. This objective shows a strong connection to the exploration-exploitation trade-off, and could also prevent the policy from getting stuck in local optima. Formally, it is given by

\begin{equation}
    J(\pi) = \mathbb{E}_{\tau \sim \rho_{\pi}}{\sum_{t=0}^{\infty} [r(s_t, a_t)}+\beta\mathcal{H}(\pi(\cdot|s_t))], 
\label{eq:merl}
\end{equation}

\noindent where $\beta$ is the temperature parameter that controls the stochasticity of the optimal policy.

\section{Problem Formulation}

\subsection{Motivation}
Given a fixed search space, GAN architecture search agents aim to discover an optimal network architecture on a given generation task. Existing RL methods update the policy network by using batches of entire architectures sampled from the current policy. Even though these data samples are only used for the current update step, the sampled GAN architectures nevertheless require  tedious training and evaluation processes. The sampling efficiency is therefore very low resulting in limited learning progress of the agents. Moreover, the entire architecture sampling leads to a high variance, which might influence the stability of the policy update.

The key motivation of the proposed methodology is to stabilize and accelerate the learning process by step-wise sampling instead of entire-trajectory-based sampling and making efficient use of past experiences from previous policies. To achieve this, we propose to formulate the GAN architecture search problem as an MDP and solve it by off-policy reinforcement learning. 

\subsection{GAN Architecture Search formulated as MDP}
We propose to formulate the GAN architecture search problem as a  Markov decision process~(MDP) which enables state-based sampling. It further boosts the learning process and overcomes the potential challenge of a large variance stemming from sampling entire architectures  that makes it inherently difficult to train a policy using off-policy data. 

Formulating GAN architecture search problem as an MDP provides a mathematical description of architecture search processes. An MDP can be described as a tuple ($S,A,r,P,\rho$), where $S$ is the set of states that is able to precisely describe the current architecture  (such as the current cell number, the structure or the performance of the architectures), $A$ is the set of actions that defines the architecture design of the next cell, $r (s,a)$ is the reward function used to define how good the architecture is, $P (s'|s,a)$ is the transition probability function indicating the training process, and  $\rho (s)$ is the initial architecture. We define a cell as an architecture block we are using to search in one step.  The design details of states, actions, and rewards is discussed in Section~\ref{sec:proposed method}. 

It is important to highlight that the  formulation proposed in this paper has two main differences compared to previous RL methods for neural architecture search. Firstly, it is essentially different to the classic RL approaches for NAS~\cite{zoph2016neural}, which formulate the task as an optimization problem over the entire trajectory/architecture. Instead, the MDP formulation proposed here enables us to do the optimization based on the disentangled steps of cell design. Secondly, it is also different to the progressive formulation used by AutoGAN~\cite{gong2019autogan}, where the optimization is based on the best performance of the current architecture level and can potentially lead to a local minimum solution. Instead, the proposed formulation enables us to potentially conduct a more global optimization using cumulative reward without the burden of calculating the reward over the full trajectory at once. It is important to point out that the multi-level optimization formulation used in AutoGAN~\cite{gong2019autogan} does not have this property.

\begin{figure}[htbp]
    \centering
    \includegraphics[scale = 0.33]{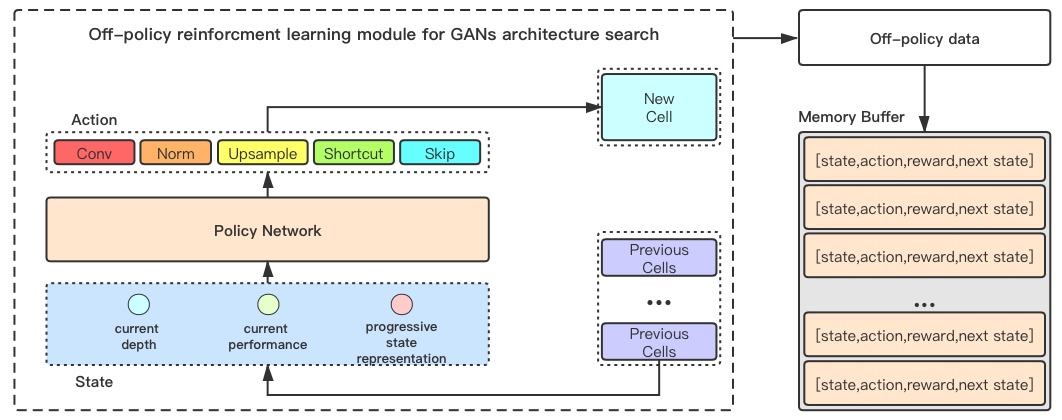}
    \caption{Overview of the proposed E$^2$GAN: the off-policy reinforcement learning module for GAN architecture search. The entire process comprises five steps: 1)The agent observes the current state $s_t$, which is designed as  s=[Depth, Performance, Progressive state representation]. 2)The agent makes a decision $a_t$ on how to design the cell added to previous cells according to the state information.  $a_t$ includes the skip options, upsampling operations, shortcut options, different types of convolution blocks and a normalization block 3)Progressively train the new architecture, obtain the reward $r_t$ and the new state $s_{t+1}$ information and then loop over it again. 4)Save the off-policy memory tuple $[s_t,a_t,r_t,s_{t+1}]$ into the memory buffer. 5)Sample a batch of data from the memory buffer to update the policy network.}
    \label{fig:main}
\end{figure}

\section{Off-policy RL for GAN Architecture Search}\label{sec:proposed method}

In this section, we integrate off-policy RL in the GAN architecture search by making use of the newly proposed MDP formulation. We introduce several innovations to address the challenges of an off-policy learning setup. 

The MDP formulation of GAN architecture search enables us to use off-policy reinforcement learning for a step-wise optimization of the entire search process to maximize the cumulative reward.

\subsection{RL for GAN Architecture Search}
Before we move on to the off-policy solver, we need to design the state, reward, and action to meet the requirements of both the GAN architecture design, as well as of the MDP formulation.
\subsubsection{State}
MDP requires a state representation that can precisely represent the current network up to the current step. Most importantly,  this state needs to be stable during training to avoid adding more variance to the training of the policy network. The stability requirement is particularly relevant since the policy network relies on it to design the next cell. The design of the state is one of the main challenges we face when adopting off-policy RL to GAN architecture search.

Inspired by the progressive GAN~\cite{karras2017progressive}, which has shown to improve generation quality in intermediate RGB outputs of each architecture cell, we propose a progressive state representation for GAN architecture search. Specifically, given a fixed batch of input noise, we adopt the average output of each cell as the progressive state representation. We down-sample this representation to impose a constant size across different cells. Note that there are alternative ways to encode the network information. For example, one could also deploy another network to encode the previous layers. However, we find the proposed design efficient and also effective. 

In addition to the progressive state representation, we also use network performance (Inception Score / FID) and layer number to provide more information about the state. To summarize, the designed state $s$ includes the depth, performance of the current architecture, and the progressive state representation.
\begin{figure}[htbp]
    \centering
    \includegraphics[scale = 0.25]{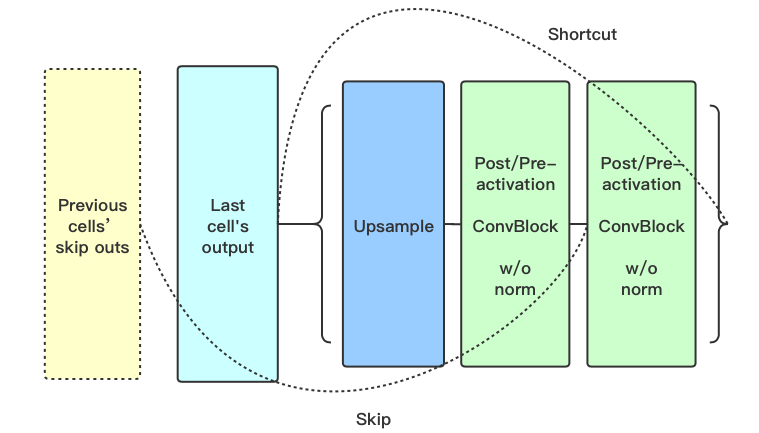}
    \caption{The search space of a generator cell in one step.  The search space is directly taken from AutoGAN~\cite{gong2019autogan}.}
    \label{fig:cell}
\end{figure}

\subsubsection{Action}
Given the current state, which encodes the information about previous layers, the policy network decides on the next action. The action describes the architecture of one cell. For example, if we follow the search space used by AutoGAN~\cite{gong2019autogan}, action will contain skip options, upsampling operations, shortcut options, different types of convolution blocks, and the normalization option, as shown in Figure~\ref{fig:cell}. 

This can then be defined as $a=[conv,norm,upsample,shortcut,skip]$. The action output by the agent will be carried out by a softmax classifier decoding into an operation. To demonstrate the effectiveness of our off-policy methods and enable a fair comparison,  in all of our experiments, we use the same search space as AutoGAN~\cite{gong2019autogan}, which means we search for generator cells, and the discriminator architecture  is pre-designed and growing as the generator becomes deeper.  More details on the search space are discussed in Section~\ref{experiments}. 
\subsubsection{Reward}
We propose to design the reward function as the \textbf{performance improvement} after adding the new cell. In this work, we use both Inception Score (IS) and Frchet Inception Distance (FID) as the indicators of the network performance. Since IS score is progressive (the higher the better) and FID score is degressive (the lower the better), the  proposed reward function can be formulated as:
\begin{equation}
    R_t(s,a) = IS(t)-IS(t-1) + \alpha(FID(t-1)-FID(t)),
\end{equation}
where $\alpha$ is a factor to balance the trade-off between the two indicators. We use $\alpha=0.01$ in our main experiments. The motivation behind using a combined reward is based on an empirical finding indicating that IS and FID are not always consistent with each other and can lead to a biased choice of architectures. A detailed discussion about the choice of indicators is provided in Section~\ref{AS}.

By employing the performance improvement in each step instead of only using performance as proposed in~\cite{gong2019autogan}, RL can maximize the expected sum of rewards over the entire trajectory. This enables us to target the potential global optimal structure with the highest reward:

\begin{equation}
   J(\pi) = \sum_{t=0} \mathbb{E}_{(s_t,a_t)\sim p(\pi)} R(s_t,a_t) =  \mathbb{E}_{architecture\sim p(\pi)} IS_{final} - \alpha FID_{final}, 
\end{equation}
where $IS_{final}$ and $FID_{final}$ are the final scores of the entire architecture.

\subsection{Off-policy RL Solver}
The proposed designs of state, reward, and action fulfill the criteria of MDPs and makes it possible to stabilize the training using off-policy samples. We are now free to choose any off-policy RL solver to improve data efficiency.

In this paper, we apply the off-the-shelf soft actor-critic algorithm (SAC)~\cite{haarnoja2018soft}, an off-policy actor-critic deep RL algorithm based on the maximum entropy reinforcement learning framework, as the learning algorithm. It has demonstrated to be 10 to 100 times more data-efficient compared to any other on-policy algorithms on traditional RL tasks. In SAC, the actor aims at maximizing expected reward while also maximizing entropy. This increases training stability significantly and improves the exploration during training. 

For the learning of the critic, the objective function is defined as:
\begin{equation}
    J(Q) = \mathbb{E}_{(s,a)\sim \mathcal{D}}\left[\frac{1}{2}(Q(s,a)-Q_{target}(s,a))^2\right]
\end{equation}
where $Q_{target}$ is the approximation target for $Q$ :

\begin{equation}
Q_{target}(s,a) = Q(s,a) + \gamma Q_{target}(s', f(\epsilon,s'))
\end{equation}
The objective function of the the policy network is given by:
\begin{equation}
\begin{aligned}
J(\pi) = \mathbb{E}_{ \mathcal{D}}\left[ \beta [\log(\pi_\theta(f_\theta(\epsilon,s)|s))]-Q(s,f_\theta(\epsilon,s)) \right]
\label{SAC}
\end{aligned}
\end{equation}
where $\pi_\theta$ is parameterized by a neural network $f_\theta$, $\epsilon$ is an input vector consisting of Gaussian noise, and the $\mathcal{D}\doteq\{(s,a,s',r)\}$ is the replay buffer  for storing the MDP tuples~\cite{mnih2015human}. $\beta$ is a positive Lagrange multiplier that controls the relative importance of the policy entropy versus the safety constraint. 
\subsection{Implementation of E$^2$GAN}
In this section, we present the implementation details of the proposed off-policy RL framework E$^2$GAN.  The training process is briefly outlined in Algorithm~\ref{algo:RSAC}.
\begin{algorithm}[tb]
   \caption{Pseudo code for E$^2$GAN  search}
   \label{algo:RSAC}
\begin{algorithmic}
    \STATE Input hyperparameters, learning rates $\alpha_{\phi_{Q}}$,$\alpha_\theta$
   \STATE Randomly initialize a Q network $Q(s, a)$ and policy network $\pi(a|s)$ with parameters $\phi_{Q}$,  $\theta$ and the Lagrange multipliers $\beta$, 
    \STATE Initialize the parameters of target networks with $\overline{\phi}_{Q}\leftarrow\phi_{Q}$,  $\overline{\theta}\leftarrow\theta$
   \FOR{each iteration}
   \STATE Reset the weight and cells of E$^2$GAN
   \FOR{each time step}
   \IF{Exploration}
   \STATE Sample $a_t$ from $\pi(s)$, add the corresponding cell to E$^2$GAN
   \ELSIF{Exploitation}
    \STATE Choose the best $a_t$ from $\pi(s)$ and add the corresponding cell to E$^2$GAN
   \ENDIF
   \STATE Progressively train the E$^2$GAN
   \STATE Observe $s_{t+1}$, $r_{t}$ and store $(s_t,a_t,r_t,s_{t+1})$ in $\mathcal{D}$
  
   \ENDFOR
   \FOR{each update step}
   \STATE Sample mini-batches of transitions from $\mathcal{D}$ and update $Q$ and $\pi$ with gradients

    \STATE Update the target networks with soft replacement:
          \begin{align}
            \overline{\phi}_{Q} &\leftarrow \tau \phi_{Q} + (1 - \tau) \overline{\phi}_{Q} \notag \\
            \overline{\theta} &\leftarrow \tau \theta +(1 - \tau) \overline{\theta} \notag
          \end{align}
   \ENDFOR
   \ENDFOR

\end{algorithmic}
\end{algorithm}


\subsubsection{Agent Training}
Since we reformulated the NAS as a multi-step MDP, our agent will make several decisions in any trajectory $\tau=[(s_1,a_1),...(s_n,a_n)]$. In each step, the agent will collect this experience $[s_t,a_t,r_t,s_{t+1}]$ in the memory buffer $\mathcal{D}$. Once the threshold of the smallest memory length is reached, the agent is updated using the Adam~\cite{kingma2014adam} optimizer via the objective function presented in Eq.~\ref{SAC} by sampling a batch of data from the memory buffer $\mathcal{D}$ in an off-policy way.

The entire search comprises two periods: the exploration period and the exploitation period. During the exploration period, the agent will sample any possible architecture. While in the exploitation period, the agent will choose the best architecture, in order to quickly stabilize the policy.

The exploration period lasts for 70$\%$ of iterations, and the exploitation takes 30$\%$ iterations. Once the memory threshold is reached, for every exploration step, the policy will be updated once. For every exploitation step, the policy will be updated 10 times in order to  converge quickly. 

\subsubsection{Proxy Task}
We use a progressive proxy task in order to collect the rewards fast. When a new cell is added, we train the current full trajectory for one epoch and calculate the reward for the current cell. Within a trajectory, the previous cells' weights will be kept and trained together with the new cell. In order to accurately estimate the Q-value of each state-action pair, we reset the weight of the neural network after finishing the entire architecture trajectory design. 

\section{Experiments} \label{experiments}

\subsection{Dataset}
In this paper, we use the CIFAR-10 dataset~\cite{krizhevsky2009learning} to evaluate the effectiveness and efficiency of the proposed E$^2$GAN framework. The CIFAR-10 dataset consists of 50,000 training images and 10,000 test images with a $32 \times 32$ resolution. We use its training set without any data augmentation technique to search for the architecture with the highest cumulative return for a GAN generator.
Furthermore, to evaluate the transferability of the discovered architecture, we also adopt the STL-10 dataset~\cite{coates2011analysis} to train the network without any other data augmentation to make a fair comparison to previous works.
\subsection{Search Space} 
To verify the effectiveness of the off-policy framework and to enable a  fair comparison, we use the same search space as used in the AutoGAN experiments~\cite{gong2019autogan}. There are five control variables: 1)Skip operation, which is a binary value indicating whether the current cell takes a skip connection from any specific cell as its input. Note that each cell could take multiple skip connections from other preceding cells. 2)Pre-activation~\cite{he2016identity} and post-activation convolution block. 3)Three types of normalization operations, including batch normalization~\cite{ioffe2015batch}, instance normalization~\cite{ulyanov2016instance}, and no normalization. 4)Upsampling operation which is standard in current image generation GAN, including bi-linear upsampling, nearest neighbor upsampling, and stride-2 deconvolution. 5)Shortcut operation.

\begin{table}[b!]
\begin{center}
\begin{tabular}{l|c|c|c}
Methods&Inception Score&FID&Search Cost (GPU days)\\
\hline
DCGAN~\cite{radford2015unsupervised}& $6.64 \pm .14$&-&$^*$\\
Improved GAN~\cite{salimans2016improved}& $6.86 \pm .06$&-&$^*$\\
LRGAN~\cite{yang2017lr}&$7.17 \pm .17$&-&$^*$\\
DFM~\cite{warde2016improving}&$7.72 \pm .13$&-&$^*$\\
ProbGAN~\cite{he2019probgan}& 7.75&24.60 &$^*$ \\
WGAN-GP, ResNet~\cite{gulrajani2017improved}& $7.86 \pm .07 $&-&$^*$ \\
Splitting GAN~\cite{grinblat2017class}&  $7.90 \pm.09 $&-&$^*$\\
SN-GAN~\cite{miyato2018spectral}&  $8.22 \pm .05 $&$21.7 \pm .01$&$^*$\\
MGAN~\cite{hoang2018mgan}&  $8.33 \pm .10 $&26.7&$^*$ \\
Dist-GAN~\cite{tran2018dist}&- &$17.61 \pm .30$&$^*$\\
Progressive GAN~\cite{karras2017progressive}&  \textbf{8.80} $\pm$ \textbf{.05 }&-&$^*$\\
Improv MMD GAN~\cite{wang2018improving} & 8.29&16.21&$^*$\\
\hline 
\hline
Random search-1~\cite{gong2019autogan}& 8.09 &17.34&-\\
Random search-2~\cite{gong2019autogan}& 7.97 &21.39&-\\
\hline
AGAN~\cite{wang2019agan}&$8.29 \pm .09 $&30.5&1200\\
AutoGAN-top1~\cite{gong2019autogan} &  $8.55 \pm .10 $ &12.42&2\\
AutoGAN-top2~\cite{gong2019autogan}& $8.42 \pm .07 $&13.67&2\\
AutoGAN-top3~\cite{gong2019autogan}   &  $8.41 \pm .11 $&13.68&2\\
\hline
E$^2$GAN-top1 & $8.51\pm .13$&\textbf{11.26}&0.3\\
E$^2$GAN-top2 & $8.50\pm .09$&12.96&0.3 \\
E$^2$GAN-top3 & $8.42\pm .11$&12.48&0.3 \\
\end{tabular}
\end{center}
\caption{Inception score and FID score of unconditional image generation task on CIFAR-10. We achieve \textbf{a highly competitive FID of 11.26} compared to published works. We mainly compare our approach with RL-based NAS approaches: AGAN~\cite{wang2019agan} and AutoGAN~\cite{gong2019autogan}. Architectures marked by ($^*$) are manually designed.}
\label{rcifar}
\end{table}


\begin{figure}[ht!bp]
    \centering
    \includegraphics[scale = 0.19]{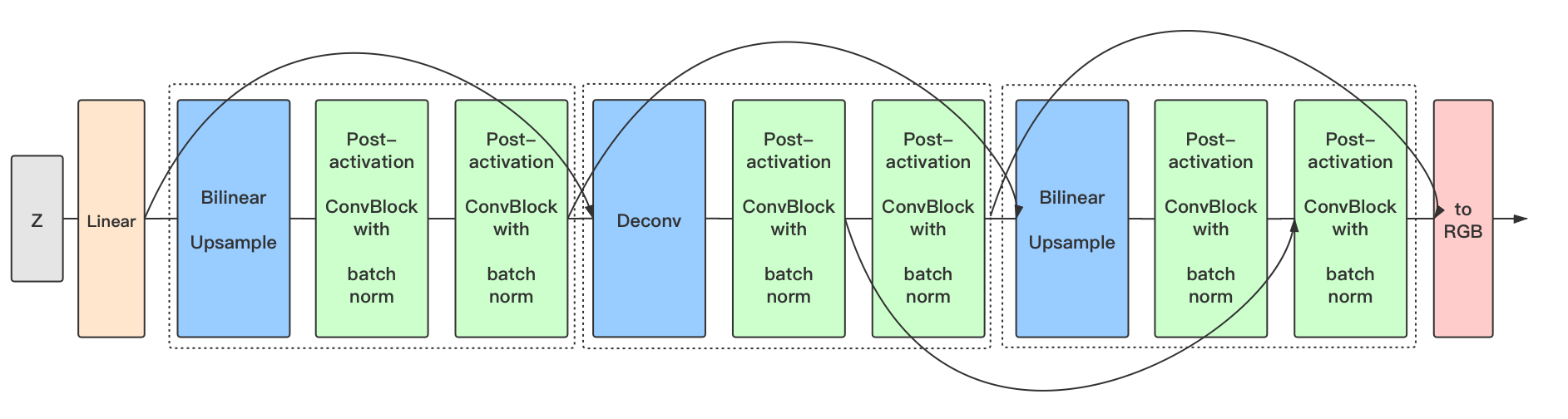}
    \caption{The generator architecture discovered by E$^2$GAN on CIFAR-10.}
    \label{fig:E$^2$GAN}
\end{figure}
\subsection{Results}
The generator architecture discovered by E$^2$GAN on the CIFAR-10 training set is displayed in Figure~\ref{fig:E$^2$GAN}. For the task of unconditional CIFAR-10 image generation (no class labels used), several notable observations could be summarized:

\begin{itemize}
\item[*] E$^2$GAN prefers post-activation convolution block to pre-activation convolution blocks. This finding is contrary to AutoGAN's preference, but coincides with previous experiences from human experts.
\item[*] E$^2$GAN prefers the use of batch normalization. This finding is also contrary to AutoGAN's choice, but is in line with experts' common practice. 
\item[*] E$^2$GAN prefers bi-linear upsample to nearest neighbour upsample. This in theory provides finer upsample ability between different cells. 
\end{itemize}

\begin{table}[b!]
\begin{center}
\begin{tabular}{l|c|c|c}
Methods&Inception Score&FID&Search Cost (GPU days)\\
\hline
D2GAN~\cite{nguyen2017dual}& 7.98&-& Manual\\
DFM~\cite{warde2016improving}&$8.51 \pm .13$&-&Manual\\
ProbGAN~\cite{he2019probgan}& $8.87 \pm .095$&46.74 &Manual \\
SN-GAN~\cite{miyato2018spectral}&  $9.10 \pm .04 $&$40.1 \pm .50$&Manual\\
Dist-GAN~\cite{tran2018dist}&- &36.19&Manual\\
Improving MMD GAN~\cite{wang2018improving}  & 9.34&37.63&Manual\\\hline
AGAN~\cite{wang2019agan}&$9.23 \pm .08$&52.7&1200\\
AutoGAN-top1~\cite{gong2019autogan}&  $9.16\pm .13 $ &31.01&2\\
\hline
E$^2$GAN-top1 & \textbf{9.51} $\pm $ \textbf{.09} &\textbf{25.35} &0.3\\

\end{tabular}
\end{center}
\caption{Inception score and FID score for the unconditional image generation task on STL-10. E$^2$GAN  uses the discovered architecture on CIFAR-10. Performance is significantly better than other RL-based competitors.}
\label{rstl}
\end{table}

Our E$^2$GAN framework only takes about 0.3 GPU day for searching while the AGAN spends 1200 GPU days and AutoGAN spends 2 GPU days.

We train the discovered E$^2$GAN from scratch for 500 epochs and summarize the IS and FID scores in Table~\ref{rcifar}.  On the CIFAR-10 dataset, our model achieves a  \textbf{highly competitive FID 11.26} compared to published results by AutoGAN~\cite{gong2019autogan}, and hand-crafted GAN~\cite{radford2015unsupervised,salimans2016improved,yang2017lr,warde2016improving,he2019probgan,gulrajani2017improved,grinblat2017class,miyato2018spectral,hoang2018mgan,wang2018improving}. In terms of IS score, E$^2$GAN is also highly competitive to AutoGAN~\cite{gong2019autogan}. We additionally report the performance of the top2 and top3 architectures discovered in one search. Both have higher performance than the respective AutoGAN counterparts.

We also test the transferability of E$^2$GAN. We retrain the weights of the discovered E$^2$GAN architecture using the STL-10 training and unlabeled set for the unconditional image generation task. \textbf{E$^2$GAN achieves a highly-competitive performance on both IS (9.51) and FID (25.35)}, as shown in Table~\ref{rstl}.

Because our main contribution is the new formulation and using off-policy RL for GAN architecture framework, we compare the proposed method directly with existing RL-based algorithms. We use the exact same searching space as AutoGAN, which does not include the search for a discriminator. As GAN training is an interactive procedure between generator and discriminator, one might expect better performance if the search is conducted on both networks. We report our scores using the exact same evaluation procedure provided by the authors of AutoGAN. The reported scores are based on the best models achieved during training on a 20 epoch evaluation interval. Mean and standard deviation of the IS score are calculated based on the 10-fold evaluation on 50,000 generated images. We additionally report the performance curve against training steps of E$^2$GAN and AutoGAN for three runs in the supplementary material.

\begin{figure}[t]
\centering
\includegraphics[width=0.35\columnwidth]{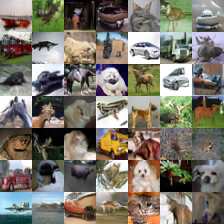}
\includegraphics[width=0.35\columnwidth]{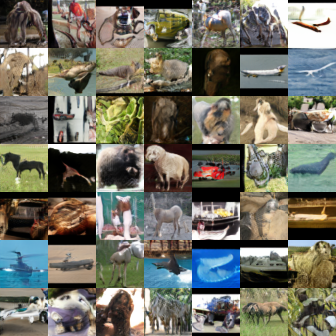}%
\caption{The generated CIFAR-10(left) and STL-10(right) results of E$^2$GAN which are randomly sampled without cherry-picking.}
\label{fig4}
\end{figure}


\begin{table}[b!]
\begin{center}
\begin{tabular}{l|c|c|c}
Methods&Inception Score&FID&Search Cost (GPU days)\\
\hline
AutoGAN-top1~\cite{gong2019autogan}&  $8.55\pm .1 $ &12.42&2\\
\hline
E$^2$GAN(IS and FID as reward)  & $8.51 \pm .13$&$11.26$&0.3\\
\hline
E$^2$GAN(IS only as reward)  & \textbf{8.81} $\pm $ \textbf{.11} &$15.64$&0.1\\

\end{tabular}
\end{center}
\caption{Performance on the unconditional image generation task for CIFAR-10 with different reward choices.}
\label{rdr}
\end{table}
\section{Discussion} \label{AS}
\subsection{Reward Choice: IS and FID}
IS and FID scores are two main evaluation metrics for GAN. We conduct the ablation study of using different combinations. Specifically, IS only ($\alpha=0$) and the combination of IS and FID ($\alpha=0.01$) as the reward. Our agent successfully discovered two different architectures. When only IS is used as the reward signal, the agent discovered a different architecture using only 0.1 GPU day. The searched architecture achieved state-of-the-art IS score of 8.86, as shown in Table~\ref{rdr}., but a relatively plain FID score of 15.78. This demonstrates the effectiveness of the proposed method, as we are encouraging the agent to find the architecture with a higher IS score. Interestingly, this special architecture shows that, at least in certain edge cases, the IS score and FID score may not always have a strong positive correlation. This finding motivates us to additionally use FID as part of the reward. When both IS and FID are used as the reward signal, the discovered architecture performs well in term of both 
metrics. This combined reward takes 0.3 GPU days (compared to 0.1 GPU days of IS only optimization) because of the relatively expensive cost of FID computation.

\subsection{Reproducibility}
\begin{figure}[t!]
    \centering
    \includegraphics[scale = 0.4]{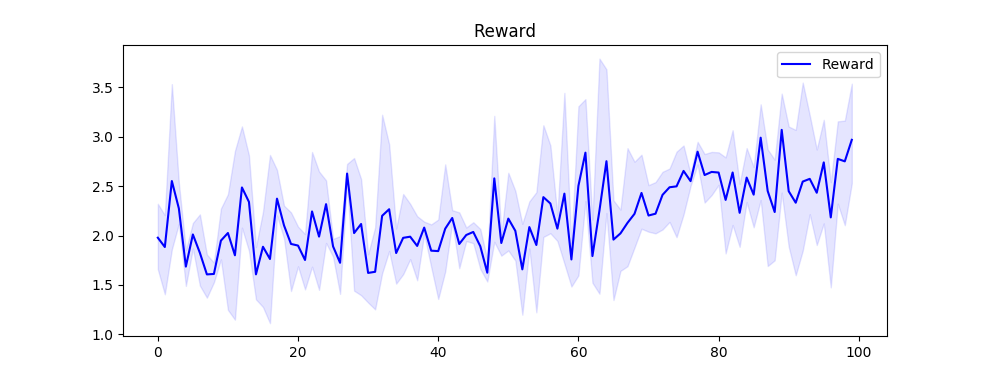}
    \caption{Training curves on architecture searching.  IS score on the proxy task against training time steps. E$^2$GAN shows relatively good stability and reproducibility.}
    \label{fig:reproducibility}
\end{figure}
We train our agent over 3 different seeds. As shown in Figure~\ref{fig:reproducibility}, we observe that our agent steadily  converged the policy in the exploitation period. E$^2$GAN can find similar architectures with relatively good performance on the proxy task. 

\section{Conclusion}
In this work, we proposed a novel off-policy reinforcement learning method, E$^2$GAN, to efficiently and effectively search for GAN architectures. We reformulated the problem as an MDP process, and overcame the challenges of using off-policy data. We first introduced a new progressive state representation. We additionally introduced a new reward function, which allowed us to target the potential global optimization in our MDP formulation. The E$^2$GAN achieves state-of-the-art efficiency in GAN architecture searching, and the discovered architecture shows highly competitive performance compared to other state-of-the-art methods. In future work, we plan to simultaneously optimize the generator and discriminator architectures in a multi-agent context.

\clearpage
\section*{Acknowledgement}
The contributions of Yuan Tian, Qin Wang, and Olga Fink were funded by the Swiss National Science Foundation (SNSF) Grant no. PP00P2\_176878.

\bibliographystyle{splncs04}
\bibliography{egbib}
\end{document}